\documentclass{article}
\usepackage{ijcai22}

\usepackage{times}
\usepackage{soul}
\usepackage{url}
\usepackage[hidelinks]{hyperref}
\usepackage[utf8]{inputenc}
\usepackage[small]{caption}
\usepackage{graphicx}
\usepackage{amsmath}
\usepackage{amsthm}
\usepackage{booktabs}
\usepackage{algorithm}
\usepackage{algorithmic}
\usepackage{amsfonts}
\title{StyleCLIPDraw: Coupling Content and Style in Text-to-Drawing Translation}

\author{
Peter Schaldenbrand$^1$\footnote{Contact Author}\and
Zhixuan Liu$^2$\and
Jean Oh$^1$
\affiliations
$^1$Robotics Institute, Carnegie Mellon University\\
$^2$School of Data Science, The Chinese University of Hong Kong\\
\emails
pschalde@andrew.cmu.edu,
zhixuanliu@cuhk.edu.cn,
jeanoh@cmu.edu
}

\begin{document}

\maketitle

\begin{abstract}
    Generating images that fit a given text description using machine learning has improved greatly with the release of technologies such as the CLIP image-text encoder model; however, current methods lack artistic control of the style of image to be generated.  
    We present an approach for generating styled drawings for a given text description where a user can specify a desired drawing style using a sample image. 
    Inspired by a theory in art that style and content are generally inseparable during the creative process, 
    we propose 
    a coupled approach, known here as StyleCLIPDraw, whereby the drawing is generated by optimizing for style and content simultaneously throughout the process as opposed to applying style transfer after creating content in a sequence. 
    Based on human evaluation, 
    the styles of images generated by StyleCLIPDraw are strongly preferred to those by the sequential approach.
    Although the quality of content generation degrades for certain styles, overall considering both content \textit{and} style, StyleCLIPDraw is found far more preferred, indicating the importance of style, look, and feel of machine generated images to people as well as indicating that style is coupled in the drawing process itself.
    Our code\footnote{\url{https://github.com/pschaldenbrand/StyleCLIPDraw}}, a demonstration\footnote{\url{https://replicate.com/pschaldenbrand/style-clip-draw}}, and style evaluation data\footnote{\url{https://www.kaggle.com/pittsburghskeet/drawings-with-style-evaluation-styleclipdraw}} are publicly available.
\end{abstract}

\section{Introduction}

\begin{figure}[t]
    \centering
    \includegraphics[width=\columnwidth]{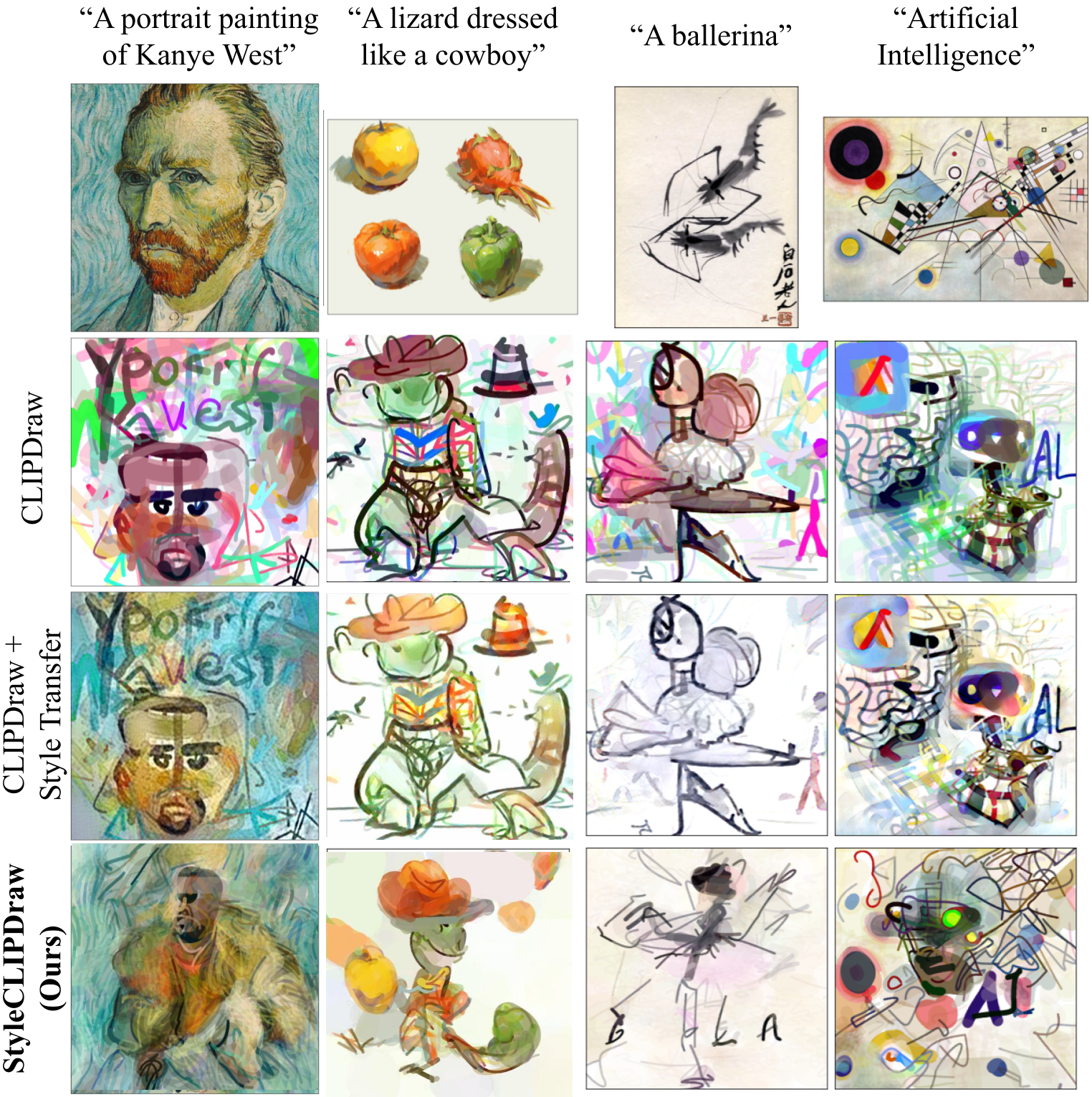}\vspace{-5pt}%
    \caption{Comparison between our (StyleCLIPDraw) method and the baseline (CLIPDraw + Style Transfer). The top row shows the language input followed by the style image input.}\vspace{-15pt}%
    \label{fig:style_transfer_results}%
\end{figure}

\begin{figure*}
    \centering
    \includegraphics[width=\textwidth]{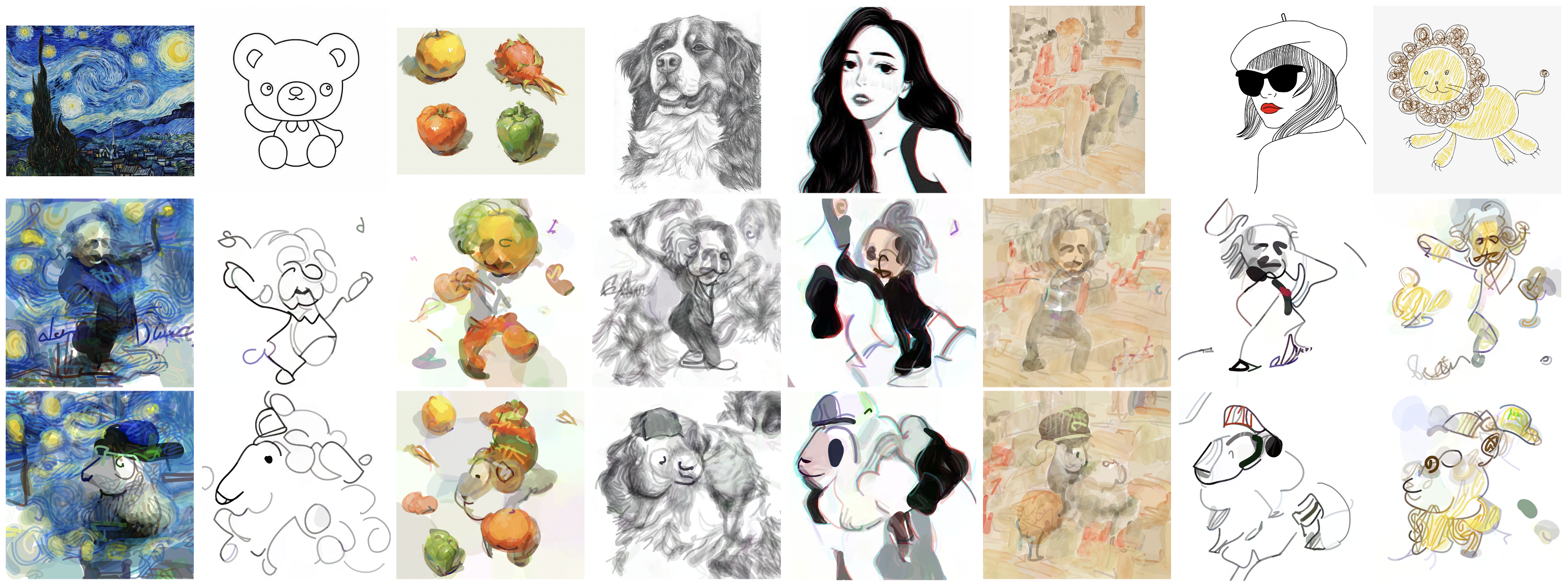}\vspace{-5pt}%
    \caption{The top row shows the style image used to generate the following images.  The second row used the text prompt ``Albert Einstein dancing" and the last row, ``A sheep wearing a top hat."}\vspace{-12pt}%
    \label{fig:main_results}%
\end{figure*}

Creating visual content is a challenging task that requires skill, time, and resources.  Painting lessons can be expensive and prohibitively time consuming for many people to participate in.  While visual content creation skills may be a privilege, there are common skills that the majority of people can use for sharing their ideas, such as verbal communication.  For instance, people may find it difficult to draw a complex scene, but they are likely able to describe it using language.  In this context, our research studies how Artificial Intelligence (AI) can act as a tool to overcome this barrier and empower people to appreciate the joy of creative practices that they otherwise would be incapable of. In this paper, we specifically focus on generating drawings from text description. 

\begin{figure}[t]
    \centering
    \includegraphics[width=\columnwidth]{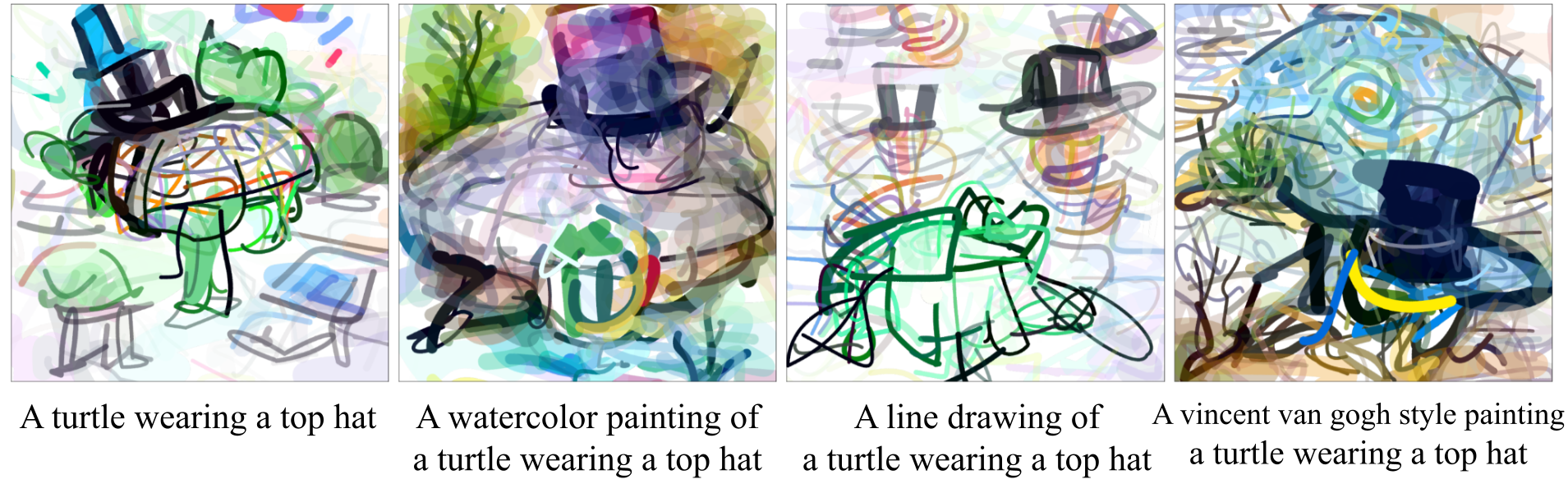}\vspace{-5pt}%
    \caption{Controlling the style of CLIPDraw generated drawings by altering the text description.}\vspace{-10pt}%
    \label{fig:clipdraw_text_style}%
\end{figure}

Language is a common form of communication to describe visual information at some level of abstraction.  Recent technological advances such as OpenAI's Contrastive Language-Image Pre-training (CLIP) \cite{radford2021-clip} have enabled high quality language-vision translation methods such as Dall-E \cite{ramesh2021-dalle}, GLIDE \cite{nichol2021-glide}, and CLIPDraw \cite{frans2021-clipdraw}.  
Using these technologies, people can write short, language descriptions to create an image in collaboration with an AI system; however, as shown in the drawings produced by CLIPDraw, e.g., the second row of Figure~\ref{fig:style_transfer_results}, all have a similar visual style.   
The language description can be used to alter the style of the drawing but only to a small extent, as shown in Figure \ref{fig:clipdraw_text_style}, since it is not trivial to describe a drawing style in words. 


Our approach is based on the assumption that adding more control to visual content creation systems is a way to enable users of the systems to feel that they have artistic ownership and control of the output. 
It has been reported that humans, in order to feel independent and take ownership, prefer having control to being served by a fully autonomous system~\cite{selvaggio2021autonomy,bretan2016-roboticDrummer}.  
If a machine performs a task for a human user without the human's input, the user may be satisfied with the product but feel that they did not participate enough in the process to feel that they own the artistic aspects of the output as in the case for professional animators~\cite{bateman2021-creatingForCreatives}. 
Furthermore, retaining input in the task can be a way to assert a person's humanity and autonomy when they have been limited either physically or mentally \cite{kim2011autonomy}. 
In this context, we propose an approach where a user can specify a style using an example image in addition to text descriptions. 



The task of machine style transfer has long been studied \cite{gatys2016image,kolkin2019-strots} where a style is applied to a given content image. This leads to an intuitive decoupled approach using existing technologies in succession, e.g., CLIPDraw generates the drawing from text, then a style transfer algorithm styles the drawing using a given style image.  
While this method does combine style and content, the combination is decoupled. 

In art, style (which is often referred to as ``form") and content are generally linked.  
Style and content could be applied agnostic to each other, but generally one informs the other. 
\cite{robertson1967-formAndContent} explains that the same content represented with different styles creates distinct artworks with varying messages or meanings, and that only a small, abstract resemblance of the content exists between each of the different styled artworks; therefore, the content, meaning, and message of an artwork is contingent on the style choices for that work.
Following this theory, 
our approach, StyleCLIPDraw, generates drawings using language and style as input with the prior knowledge that style and content are coupled in the drawing process. 

This paper includes the following contributions: 1) we propose a coupled text-and-style-to-image generation approach; 
2) we release our full human evaluation results and 352 AI-generated drawings publicly as the first dataset of its kind for evaluating the style of images thoroughly \url{https://www.kaggle.com/pittsburghskeet/drawings-with-style-evaluation-styleclipdraw};
%
3) we open source StyleCLIPDraw with a public Google Colab notebook that allows people to use it easily for free \url{https://github.com/pschaldenbrand/StyleCLIPDraw}; 4) An online demonstration that does not require programming knowledge is available at \url{https://replicate.com/pschaldenbrand/style-clip-draw}.

\begin{figure*}[t]
    \centering
    \includegraphics[width=\textwidth]{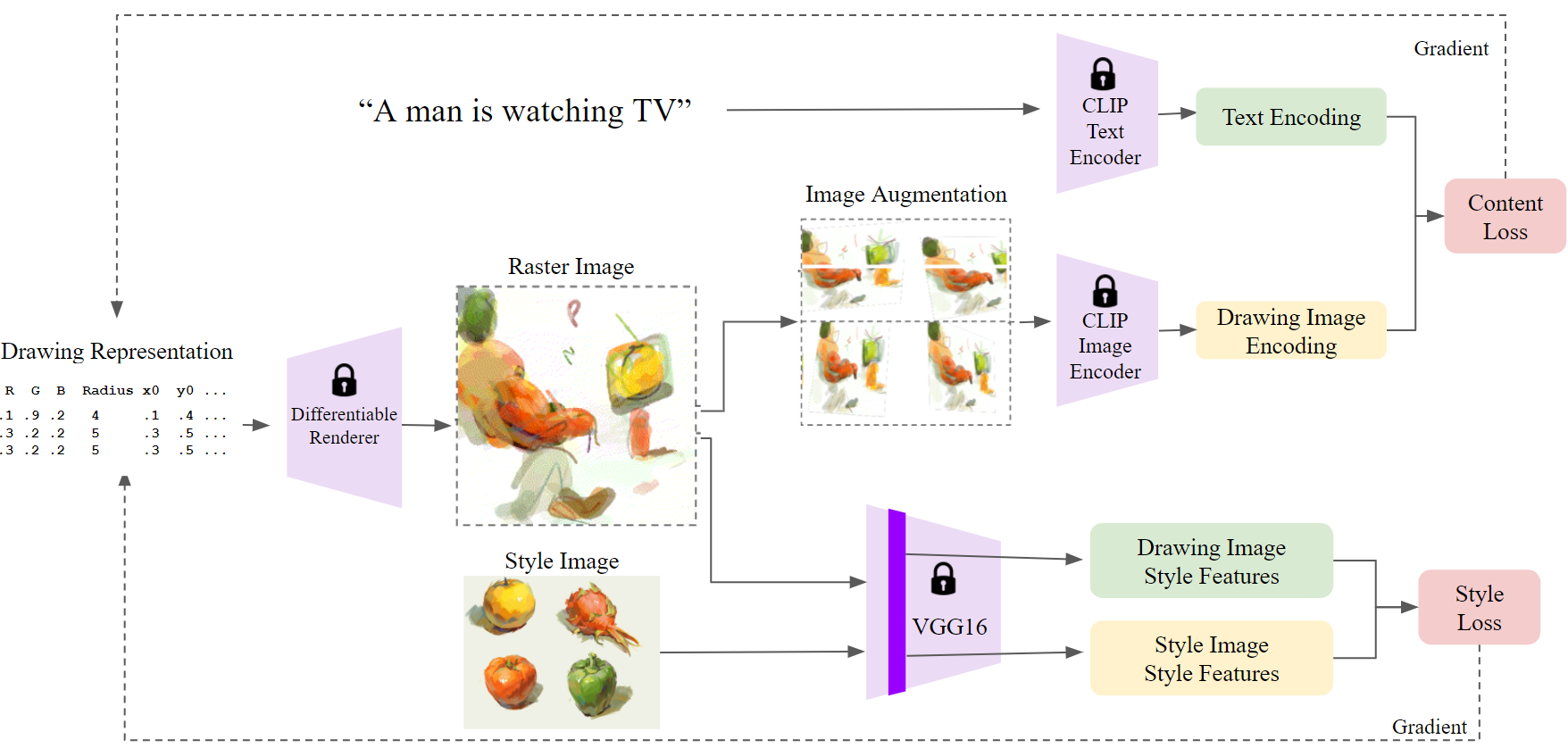}\vspace{-5pt}%
    \caption{StyleCLIPDraw optimizes a drawing representation by computing two losses: one for content using the text description and the other for style using a style image.  The drawing representation is rasterized, then style and content features are extracted using CLIP and VGG16 models respectively.  Style and content features are compared using distance functions to compute a loss.}\vspace{-12pt}%
    \label{fig:method_diagram}%
\end{figure*}%

\section{Related Work}

\subsection{Text-to-Image Synthesis Models} 

The most general and successful methods for text-to-image synthesis recently either involve training a model in a supervised fashion to perform the task or using a pre-trained image generation model with CLIP \cite{radford2021-clip} to guide generation.  Examples of the former include Dall-E \cite{ramesh2021-dalle} and GLIDE \cite{nichol2021-glide} which were trained using roughly 400 million image-text pairs.  Training these models on such a huge dataset took an incredible amount of resources.  Neither Dall-E nor GLIDE have been released publicly in full.

\subsection{CLIP-Guided Text-to-Image Synthesis}

A more computationally efficient method of text-to-image synthesis is creating one text-to-image translation at a time and utilizing pre-trained models.  Any pretrained image generation model can be used as the image generator.  A latent vector used as input to the generator model is randomly initialized and acts as the parameter space to be optimized.  
CLIP encodes the image generated from the latent space as well as the text description given by a user.  The latent space is optimized such that the cosine distance between the generated image's encoding and the text encoding is minimized.
This method is very popular in the technological art community now and has been implemented many times with different pre-trained models \cite{Galatolo2021-ClipStyleGAn2,smith2021-clipGuided}.

\subsection{Text-to-Drawing Synthesis}

Previously mentioned methods generate raster images directly.  In this work, we focus on drawings which are represented by strokes that can be rendered into raster images.  Prior work with drawings often focuses on decomposing images into brush strokes, referred to as stroke-based rendering (SBR). 
More recent methods of SBR train deep reinforcement learning or transformer models to generate the brush strokes in a probabilistic fashion \cite{liu2021-paintTransformer,schaldenbrand2020content,huang2019-learningToPaint}.

CLIPDraw is a method that combines the action space of SBR and the methodology of CLIP-guided image generation. Rather than optimizing a model to perform every text-to-drawing synthesis task imaginable, each translation is optimized separately.  So instead of modifying the parameters of a neural network model, the the brush strokes themselves are modified to optimize the objective.

\subsection{Drawing and Style Evaluation}

In related work to altering the style of images such as style transfer \cite{gatys2016image,kolkin2019-strots}, style is not properly defined or it is defined only as relating to the color and texture of an image leaving out important concepts such as shape and space.
Artists usually use art features, such as color, line, texture, and shape \cite{thorson2020-elementsOfArt}, to describe artworks. 
In the computer vision area, research shows that utilizing these features to describe images could positively affect image retrieval \cite{alsmadi2020content} and image classification \cite{banerji2013new}. These features can be used as strong representatives of the style of an image, therefore some are used as evaluation methods for artworks \cite{kim2022formal,kim2010computer}. 
In this work, we define the style of a drawing as a combination of the 7 elements of art drawn from \cite{thorson2020-elementsOfArt} and \cite{beck2014-elementsOfArt2} which are each described in Table \ref{tab:elements_of_art}.  Note that Form is not relevant to drawings as drawings are two dimensional images.



\section{Method}

Our approach is depicted in Figure \ref{fig:method_diagram}. Two losses are computed to optimize for both content and style in a coupled manner at each iteration.  We build on the drawing representation and the content loss methodology of CLIPDraw~\cite{frans2021-clipdraw} for which we include brief descriptions for self-containment. 

\subsection{Drawing Representation}

Drawings are represented using a series of virtual brush strokes which each have parameters for a trajectory, color, and width as was defined in \cite{frans2021-clipdraw}.  The trajectories are described by cubic B\'ezier curves consisting of four control points on a two dimensional coordinate plane.  
The color of the stroke is represented by four real numbers defining the amount of red, green, and blue in the color along with an alpha channel controlling the opacity of the stroke. The widths of the brush strokes are represented by the radius of the brush stroke given in pixel units. 



The drawing representation cannot be utilized directly by models like CLIP or VGG16 which require raster images as input.  To rasterize the drawing representation, we employ the DiffVG \cite{li2020-diffVG} package which renders vector graphics in a differentiable manner.  

\subsection{Drawing Augmentation}

During optimization, minor flaws in the pre-trained model can be exploited, resulting in local minima that have high scores but poor results that are sometimes referred to as adversarial examples \cite{rebuffi2021data}. To stabilize optimization, the rasterized drawings are augmented with random crops and perspectives prior to being fed to CLIP as in \cite{frans2021-clipdraw,tian2021-es-strategies-drawing}.



\subsection{Content Loss}

The content loss is computed following the methodology of \cite{frans2021-clipdraw}. 
The input text description is the primary control for the content of the image.  
CLIP is used to connect the text description to the drawing representation.
The drawing representation ($x$) is rasterized, augmented, and then encoded using CLIP's image encoder (Eq. \ref{eq:encode_image}).  The given text description is encoded by CLIP's text encoder.  These two encodings are compared using cosine similarity ($S_C$ in Eq. \ref{eq:content_loss}) and averaged to form the loss function for the model which guides the drawings to resemble the text description. We multiply the cosine similarity by $-1$ because our optimization algorithms are designed to decrease the loss.

\begin{align}
\begin{split}
    Drawing &= \texttt{DiffVG}(x) \\
    Enc_{drawing} &= \texttt{CLIP}(\texttt{Augment}(Drawing))  \\
    Enc_{text} &= \texttt{CLIP}(text) \label{eq:encode_image}
\end{split}
\end{align}
\begin{align}
    \mathcal{L}_{content} &= - S_C(Enc_{drawing}, Enc_{text}) \label{eq:content_loss}
\end{align}

\subsection{Style Loss}

The style of the generated drawing is specified through a given example style image.
The style of the example image and generated drawing are made up of the texture, use of space, shapes of objects, lines, colors, and value of the colors in the images.
To extract this style information from raster images, it is common to use pretrained neural networks \cite{gatys2016image,kolkin2019-strots}.  
We extract style features based on the approach from STROTSS \cite{kolkin2019-strots}.
Features are extracted from early layers of the VGG16 object detection model from both the drawing and the style image.  
Early layers of the model tend to extract low level image features such as texture and color \cite{chandrarathne2020-cnnFeatures}, so these features are known to correlate well with the style information that humans would agree on.
We compare these features with Earth Mover's Distance to compute a loss term for style which can be back propagated.

\subsection{Optimization Algorithm}

The drawing representation is randomly initialized to begin the optimization algorithm with a controllable number of brush strokes. 
As seen in Figure \ref{fig:method_diagram}, the drawing is rasterized and used to compute losses for content and style.  
The objective of the optimization algorithm is to find the drawing representation ($\hat{x}$ in Eq. \ref{eq:objective}) such that the weighted sum of the content ($\mathcal{L}_{content}$) and style ($\mathcal{L}_{style}$) losses are minimized.  The weights, $\lambda_1$ and $\lambda_2$, control how strong the text or style image should influence the appearance of the drawing. We report on the impact of these two weights in Section~\ref{sec:results-weights}.

\begin{align}
    \hat{x} &= \min_{x} [ \lambda_1 \mathcal{L}_{content} + \lambda_2 \mathcal{L}_{style} ] \label{eq:objective}
\end{align}

In every iteration of the optimization algorithm, the losses are computed, the derivatives are calculated with backpropagation, and the drawing representation is altered to decrease the losses.  The modification of the drawing representation is performed by the RMSProp algorithm.  There is a separate instance of RMSProp for the stroke trajectories,  stroke widths, and stroke colors since different learning rates (e.g., 0.3, 0.3, and 0.03 respectively) need to be used because of the varying magnitudes of the data.

The two loss terms are combined and optimized in one concerted backpropagation step as they are added together in Eq. \ref{eq:objective}.  
Alternatively, each loss can be optimized separately in an alternated fashion where content is optimized for some number of iterations followed by style optimized for several iterations.  We call these approaches ``concerted optimization" and ``alternated optimization" respectively and discuss their efficiency in Section~\ref{sec:results-opt}.


\section{Experiments}

\textbf{Settings:} All experiments were conducted using our publicly available notebooks on Google Colab.

\subsection{Decoupled Baseline}\label{sec:baseline}

Styled text-to-drawing synthesis can be conducted using existing technologies chained together in sequence.  This formed the baseline in this work: The language input is converted into a drawing using CLIPDraw, then the drawing is styled with the style image using the STROTSS style transfer algorithm as a post-process.



\subsection{Human Evaluation Survey}\label{sec:survey}

\textbf{Main experiment:} We hand crafted\footnote{On a pilot study, prompts from the Flickr image caption dataset~\cite{plummer2015flickr30k} were found to be too complex.} 22 text prompts and chose 8 different style images to generate 176 drawings for the StyleCLIPDraw and baseline methods.  For each style image and prompt pair, we generated 4 drawings and selected the one with the best cosine similarity between CLIP encodings of the drawing and text description.  We used the Amazon Mechanical Turk (AMT) platform to conduct the study.  Participants were first educated on what the different dimensions of art are and what they mean (Table \ref{tab:elements_of_art}).  The participants were then shown the style image that was used to generate the drawings, and the two drawings: one from StyleCLIPDraw and the other from the baseline in a random order.  There were 9 questions asked about each pair of drawings.  The first 6 were for each style element in Table \ref{tab:elements_of_art}, i.e., ``Which drawing is most similar to the Style Image in terms of [style element]?"  
The participants were also asked to pick which drawing overall fits the style of the style image best. Next, the text prompt used to generate the drawings was given, and the participants were asked to pick which drawing best fits the description, if neither fit at all, or if both fit it equally well. Lastly, the participants were asked which drawing (if any or both) fits the text \textit{and} the style best.

\noindent\textbf{Auxiliary questions:} To compare the realism of the style images used in the study, we asked the AMT workers to rate the 8 style images used in our study in 5 categories: very simple, simple, neutral, detailed, and very detailed.

\begin{table}
\begin{tabular}{p{.25\columnwidth}p{.64\columnwidth}}
\toprule
Element of Art & Description  \\
\midrule
Lines                        & Outlines capable of producing texture with their \\
Shapes               & General geometry and makeup of objects within the drawings. \\
Space                & Perspective and proportion between shapes and objects.   \\
Texture              & The way things look as if they might feel if touched.    \\
Value                & The Lightness or darkness of tones or colors.          \\
Color                & The hue, lightness, darkness, and intensity of color in a drawing.  \\
Form                 & The three dimensional properties of the work (not used in the evaluation because it is less relevant to 2D images). \\
\bottomrule
\end{tabular}
\caption{The seven elements of art that make up the style of an artwork.}
\label{tab:elements_of_art}
\end{table}

\section{Results}\label{sec:results}

We report on the findings from the analysis of our approach and then present the evaluation results. 

\subsection{Style and Content Weights}\label{sec:results-weights}

\begin{figure*}[t]
    \centering
    \includegraphics[width=\textwidth]{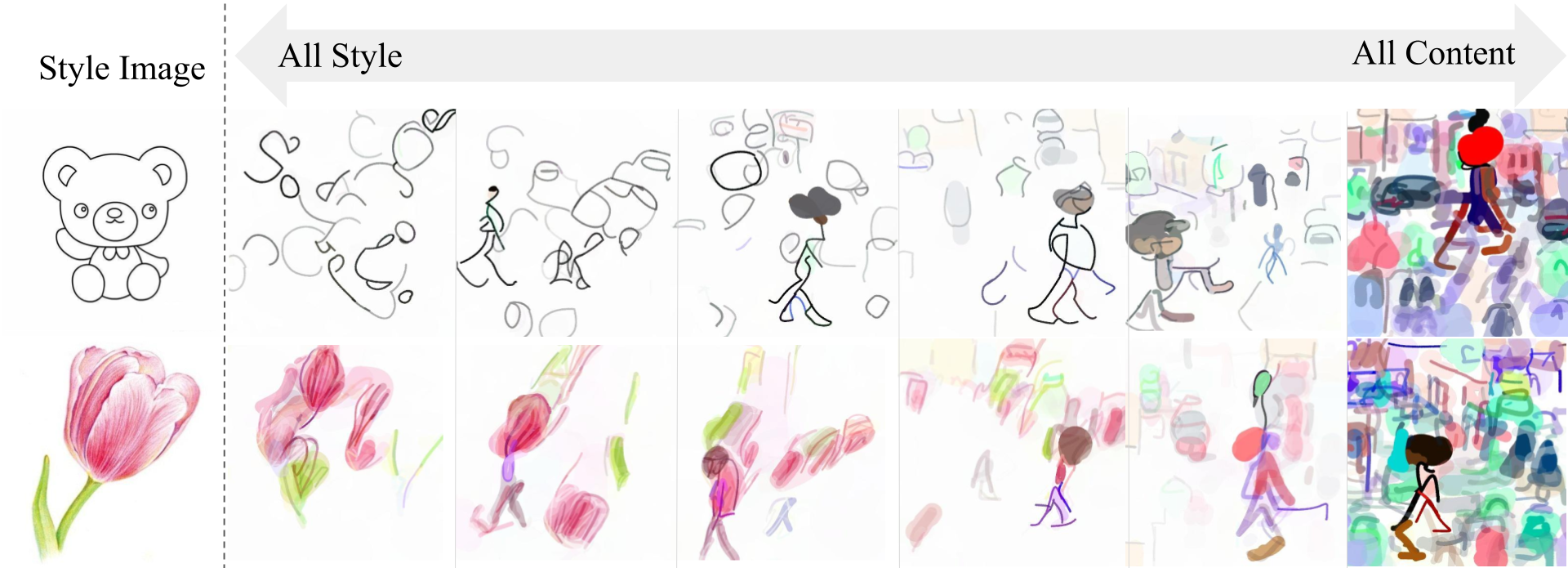}\vspace{-5pt}%
    \caption{
     From left to right: increasing the weight of the content loss and decreasing the weight of the style loss ($\lambda_1$ and $\lambda_2$ respectively in Eq. \ref{eq:objective}).
     The drawings were generated using the style image on the far left and the text prompt, ``A person is walking down a city street."}\vspace{-12pt}%
    \label{fig:style_spectrum}
\end{figure*}

We qualitatively report on how the quality of the output drawings is impacted by the choice of the style and content weights in the proposed optimization method in Equation \ref{eq:objective}. We generated drawings using the same text prompt and style image and varied the style and content weights. 
From left to right in Figure \ref{fig:style_spectrum}, drawings were generated with decreasing style loss weight and increasing content loss weight ($\lambda_2$ and $\lambda_1$ respectively in Eq. \ref{eq:objective}).  With exclusively style weight, the drawings capture the style well but lack resemblance to the text description, while with exclusively content weight the drawings represent the text well but lack style from the style image.  Using these qualitative results we chose $1.0$ for both $\lambda_1$ and $\lambda_2$ for the remainder of the paper for the balance between content and style strength.


\subsection{Concerted versus Alternated Optimization}\label{sec:results-opt}



Based on an AMT study designed similar to the main survey described in Section~\ref{sec:survey} on 51 users, the concerted and alternated optimization approaches are comparable in terms of content and style generation although a minor advantage was shown for alternated optimization in style generation; however, the concerted optimization method is drastically superior in terms of computational efficiency.  
Specifically, the runtime increase in the alternated optimization is ten fold longer than the concerted optimization and only two fold more preferred in terms of style. 
The concerted optimization approach takes about 1 minute with a P100 NVIDIA GPU on our Google Colab notebook which is promising news towards developing a real-time system.  

\begin{table}
\begin{tabular}{lcccc}
\toprule
Comparison\\Dimension         & Ours           & Baseline & Neither & Both        \\
\midrule
Lines                        & \textbf{75.4} & 24.6 & &         \\
Shapes                       & \textbf{76.6} & 23.4 & &          \\
Space                        & \textbf{82.1} & 17.9 & &          \\
Texture                      & \textbf{77.6} & 22.4 & &          \\
Value                        & \textbf{77.4} & 22.6 & &          \\
Color                        & \textbf{80.8} & 19.2 & &          \\
Style Overall                & \textbf{84.9} & 15.1 & &          \\
Content                      & 24.3          & \textbf{52.1} & 9.1 & 14.4 \\
Style \& Content & \textbf{48.1} & 23.3 & 19.8 & 8.9 \\
\bottomrule
\end{tabular}
\caption{Percentage of preference of drawings in direct comparison between our StyleCLIPDraw approach and the baseline. More details in Section \ref{sec:survey}.
}
\label{tab:direct_comparison}
\end{table}


\subsection{Qualitative Results}

In Figure \ref{fig:style_transfer_results}, we compare the performance of StyleCLIPDraw with the original CLIPDraw~\cite{frans2021-clipdraw} and the decoupled baseline approach: CLIPDraw + Style Transfer (Section ~\ref{sec:baseline}). Four prompts and style image pairs are used to generate drawings.  
On the content generation, all three approaches appear to fit the content of the given text description well. 
On the style generation, our coupled approach is able to generate the output images faithfully according to the given style input. By contrast, the decoupled baseline of applying style transfer as a postprocess to the CLIPDraw drawings alters the colors and textures of the images but lacks adaptation of other art elements 
such as shape, space, and lines.

Additional examples of how StyleCLIPDraw responds to different  prompts and style images are shown in Figure \ref{fig:main_results}. 

We note that, due to the lack of abstract style, the CLIPDraw drawings appear to capture the content of the text more clearly and literally. 
Certain styles of drawings are more constraining than others.  For instance, the more painterly style images appear to give the approach more flexibility with the output compared to the highly abstract line drawing examples. Since StyleCLIPDraw applied the style far more that the baseline, it could be true that the style hurt the appearance of the content of the image more than the baseline.
Using the auxiliary questions in our survey, we found that there is a strong, positive correlation (0.72 with a p-value of 0.04) between the divided opinion on realism of the style image and the appearance of content in the image, i.e.,
the more ambiguous the realism of the style image is, the less preferred the generated output's content is.

\subsection{Quantitative Results}

The results from the human evaluation study based on 139 participants are summarized in Table \ref{tab:direct_comparison}.  
As we speculated in the qualitative analysis, the decoupled method generated clearer contents that are easier for human users to recognize. 
On the style aspect, according to all style dimensions in Table~\ref{tab:elements_of_art}, our StyleCLIPDraw approach was vastly preferred to the decoupled approach.
This result provides strong evidence for our assumption that style and content generation is a coupled process and that the two objectives should be jointly optimized throughout the creation process.

Overall, when asked about both style \textit{and} content, the participants vastly preferred our approach over the baseline, underscoring the importance of style, look, and feel of the generated images. 
This result also provides a promising signal for our assumption that an additional control over the output can potentially improve the user's artistic ownership. 



\subsection{Limitations}
Our approach relies on existing image-text models such as CLIP that are generally trained over a gigantic amount of non-public data. Based on our experiments, the complexity of text descriptions that existing models can handle is relatively simple as seen in the examples used in the paper. 

As discussed in the results, using abstract style images seems to impede the content quality. Creating recognizable content using an abstract style generally requires more creativity. For a future direction, we consider including additional loss to also optimize for recognizability or gestalt. 

Our current method is still not fast enough to be used in real-time applications such as communication assistance. 

\section{Conclusions}
This paper addresses a question of how AI can be used to empower laypeople to engage in creative visual content generation. Building upon the assumption that adding more control by way of human input to an AI system can enable users to gain more fulfilling ownership and control over the output, we propose an approach for image generation where users can have an example-based stylistic control as well as a text-based content control.
Following the theory in art that the style of the drawing is intertwined with its content, we propose StyleCLIPDraw, a coupled approach where style and content are jointly optimized at every iteration from start to end. 

Based on 139 random human evaluators, StyleCLIPDraw was preferred to the decoupled approach $84.9$\% of time in terms of various style elements, matching the underlying theory of our approach.  Notably, despite the fact that it was less preferred in content solely, when considering both style \textit{and} content, StyleCLIPDraw drawings were highly preferred. This result is an indication that style is as much as or even more important than the content of the image to people's perceptions of generated images.  

As part of contributions, we open source our approach to promote future research on this direction.  Our online demonstration provides a tool for non-technical people to also benefit from our work. We also make the results from the study available publicly that include 352 digital drawings with comparisons across style as a whole, individual elements of style, and content.  To our knowledge, this is the first public dataset with information about the quality of drawings with respect to the individual elements of art and can contribute to future automatic methods of evaluating the style of generated images.
Future work on StyleCLIPDraw includes adding more artistic control in the form of compositional input and improving the runtime to near real-time.

\section*{Acknowledgement}
We would like to thank Benjamin Stoler for his input to this work.

\bibliographystyle{named}
\bibliography{ijcai22}

\end{document}